\documentclass[11pt,letterpaper]{article}
\usepackage{emnlp2017}
\usepackage{times}
\usepackage{latexsym}
\usepackage{amssymb}
\usepackage{amsmath}
\usepackage{graphicx}
\usepackage{caption}
\usepackage{subcaption}
\usepackage{multirow}
\usepackage{booktabs}
\usepackage{tikz}
\usepackage{xcolor}
\usepackage{stfloats}

\emnlpfinalcopy

\title{Improving Opinion-Target Extraction with Character-Level Word Embeddings}

\author{Soufian Jebbara \and Philipp Cimiano \\
  Semantic Computing Group, Bielefeld University \\
  {\tt \{sjebbara, cimiano\}@cit-ec.uni-bielefeld.de}}

\date{}

\graphicspath{{images/}}

\tikzstyle{every picture}+=[remember picture]
\tikzstyle{none} = [shape=rectangle,inner sep=2pt,outer sep=1pt,text depth=0pt]
\tikzstyle{sentiment} = [shape=rectangle,inner sep=2pt,outer sep=1pt,text depth=7pt]
\tikzstyle{aspect} = [draw,shape=rectangle,inner sep=2pt,outer sep=1pt,text depth=0pt,fill=white!70!blue]
\tikzstyle{opinion} = [draw,dashed,line width=0.8pt,shape=rectangle,inner sep=2pt,outer sep=1pt,text depth=0pt,fill=white!40!red]
\tikzstyle{comment} = [shape=rectangle,inner sep=2pt,outer sep=1pt,text depth=0pt,scale=0.7]

\newcommand{\wrd}{{wrd}}
\newcommand{\chr}{{chr}}

\begin{document}

\maketitle

\begin{abstract}
Fine-grained sentiment analysis is receiving increasing attention in recent years.
Extracting opinion target expressions (OTE) in reviews is often an important step in fine-grained, aspect-based sentiment analysis.
Retrieving this information from user-generated text, however, can be difficult.
Customer reviews, for instance, are prone to contain misspelled words and are difficult to process due to their domain-specific language.
In this work, we investigate whether character-level models can improve the performance for the identification of opinion target expressions.
We integrate information about the character structure of a word into a sequence labeling system using character-level word embeddings and show their positive impact on the system´s performance.
Specifically, we obtain an increase by 3.3 points F1-score with respect to our baseline model.
In further experiments, we reveal encoded character patterns of the learned embeddings and give a nuanced view of the performance differences of both models.
\end{abstract}

\section{Introduction}
In recent years, there has been an increased interest in developing sentiment analysis models that predict sentiment at a more fine-grained level than at the level of a complete document.
A key task within fine-grained sentiment analysis consists in identifying so called opinion target expressions (OTE).
These are the objects of a sentiment expression. 
Consider the following example:

\vspace{3mm}
\noindent
{%
\small
\textit{%
\tikz\node[none]{``};\tikz\node[aspect](a1){Moules}; \tikz\node[none]{were}; \tikz\node[opinion](s1){excellent};\tikz\node[none]{,}; \tikz\node[aspect](a2){lobster ravioli}; \tikz\node[none]{was}; \tikz\node[none]{VERY}; \tikz\node[opinion](s2){salty};\tikz\node[none]{!};\tikz\node[none]{''};%
}%
}
\begin{tikzpicture}[overlay]
  \path[->,black,semithick](s1) edge [out=160, in=20] (a1);
  \path[->,black,semithick](s2) edge [out=340, in=210] (a2);
\end{tikzpicture}
\vspace{5mm}

\noindent
where blue boxes mark opinion targets, (dashed) red boxes the opinion terms and arrows the respective relations.
In this example, there are two sentiment statements, one positive and one negative.
The positive one is indicated by the word \textit{excellent} and is expressed towards the \textit{Moules}.
The second, negative, sentiment is indicated by the work \textit{salty} and is expressed towards the \textit{lobster ravioli}.

In this work, we consider the task of identifying such opinion target expressions in reviews as a sequence labeling problem.
A particular challenge involved in OTE identification stems from the fact that online reviews can be of low quality and contain misspelled words, novel word creations, rare words etc.
We thus hypothesize that including character-embeddings might be beneficial in the context of OTE extraction, allowing a model to be robust to spelling errors as well as generalize to unseen words.
A further challenge is that an OTE can span multiple tokens.

In this work, we thus investigate whether a character-based approach is capable of using the additional low-level information to improve upon a standard word-based baseline.
We hypothesize that character-level word embeddings capture relevant information for opinion target expression extraction that regular (skip-gram) word embeddings lack.
We propose a neural network model that learns and utilizes character-level word embeddings to extract opinion target expressions and examine its characteristics.
Our experimental analysis shows that with an increase of 3.3 points F score, the character information is indeed valuable for the task.
Further experiments reveal encoded character patterns of the learned embeddings and give a nuanced view of the performance differences of both models.

The rest of the paper is structured as follows:
Section \ref{sec:related} discusses related work from two domains: fine-grained sentiment analysis and character-level neural text processing.
In Section \ref{sec:models}, we describe our approach to address opinion target extraction and present the recurrent neural network models that we use to measure the impact of character information on the task.
We carry out our evaluation and analysis in Section \ref{sec:experiments} and examine the learned character-level word embeddings in more detail.
Finally, Section \ref{sec:conclusion} summarizes our findings and presents directions for future work.

\section{Related Work}
\label{sec:related}
Our work brings together the domains of fine-grained sentiment analysis on the one side and character-level neural text processing on the other side.
In this Section we give a brief overview of both domains and point out parallels to previous work.

\paragraph{Fine-Grained Sentiment Analysis}
\citet{agerri2015elixa} present a system that addresses opinion target extraction as a sequence labeling problem based on a perceptron algorithm with local features.

\citet{Toh2014} propose a Conditional Random Field (CRF) as a sequence labeling model that includes a variety of features such as Part-of-Speech (POS) tags and dependencies, word clusters and WordNet taxonomies.

\citet{Jakob2010} follow a very similar approach that addresses opinion target extraction as a sequence labeling problem using CRFs.
Their approach includes features derived from words, POS tags and dependency paths, and performs well in a single and cross-domain setting.

\citet{Klinger2013a,Klinger2013b} model the task of joint aspect and opinion term extraction using probabilistic graphical models and rely on Markov Chain Monte Carlo methods for inference.
They demonstrate the impact of a joint architecture on the task with a strong impact on the extraction of aspect terms, but less so for the extraction of opinion terms.

\paragraph{Character-Level Neural Network Models}
Character-level neural network models are gaining interest in many research areas such as language modeling \cite{KimJSR16}, spelling correction \cite{SakaguchiDPD17}, text classification \cite{zhang2015character} and more.
Most similar works from the area of character-level word representations can be found in \cite{santos2014learning,Santos2015,ma2016end}.
In these works, word and character level representations are successfully learned and combined to improve Part-of-Speech (POS) tagging and Named Entity Recognition (NER).

\citet{santos2014learning} and \citet{Santos2015} apply a convolutional neural network (CNN) to the raw character sequence that detects character patterns and represents them as a fixed-sized embedding vector.
The concatenated sequence of word and character-level embeddings is then used to predict POS or NER tags for each word.

\citet{ma2016end} use a similar CNN-based word structure model. However, the subsequent processing of the embedded word sequence is carried out using a bidirectional Long Short-Term Memory network (LSTM).

An example of character-level text classification not requiring any tokenization is given by \citet{zhang2015character}.
In their work, the authors perform text classification using character-level CNNs on very large datasets and obtain comparable results to traditional models based on words.
Their findings suggest that the obligatory tokenization of text is indeed something to be reconsidered.

\section{Model}
\label{sec:models}
In this work, we approach the task of extracting opinion target expressions by phrasing it as a sequence labeling problem.
Doing so allows us to extract an arbitrary number of multi-word expressions in a given text.
We use the \textbf{IOB} scheme \cite{tksveenstra99eacl} to represent OTEs as a sequence of tags.
According to this scheme, each word in our text receives one of 3 tags, namely \textbf{I}, \textbf{O} or \textbf{B} that indicate if the word is at the \textbf{B}eginning\footnote{Note that the \textbf{B} token is only used to indicate the boundary of two consecutive phrases.}, \textbf{I}nside or \textbf{O}utside of an expression:

\vspace{5mm}
{
    \small
    \begin{tabular}{cccccccc}
        \textit{The} & \textit{wine} & \textit{list} & \textit{is} & \textit{also} & \textit{really} & \textit{nice} & \textit{.} \\
        O & \textbf{I} & \textbf{I} & O & O & O & O & O 
    \end{tabular}
}
\vspace{5mm}

The task is thus reduced to mapping a sequence of words to a sequence of tags.
We model the sequence labeling task using recurrent neural networks (RNN).
RNNs allow us to easily integrate character-level knowledge into the model in the form of character-level word embeddings.
To quantify the impact of these embeddings, we compare it to a baseline model that only uses word level embeddings.

\subsection{Baseline Model}
\label{sec:baselinemodel}
\begin{figure}
    \centering
    \includegraphics[width=\columnwidth]{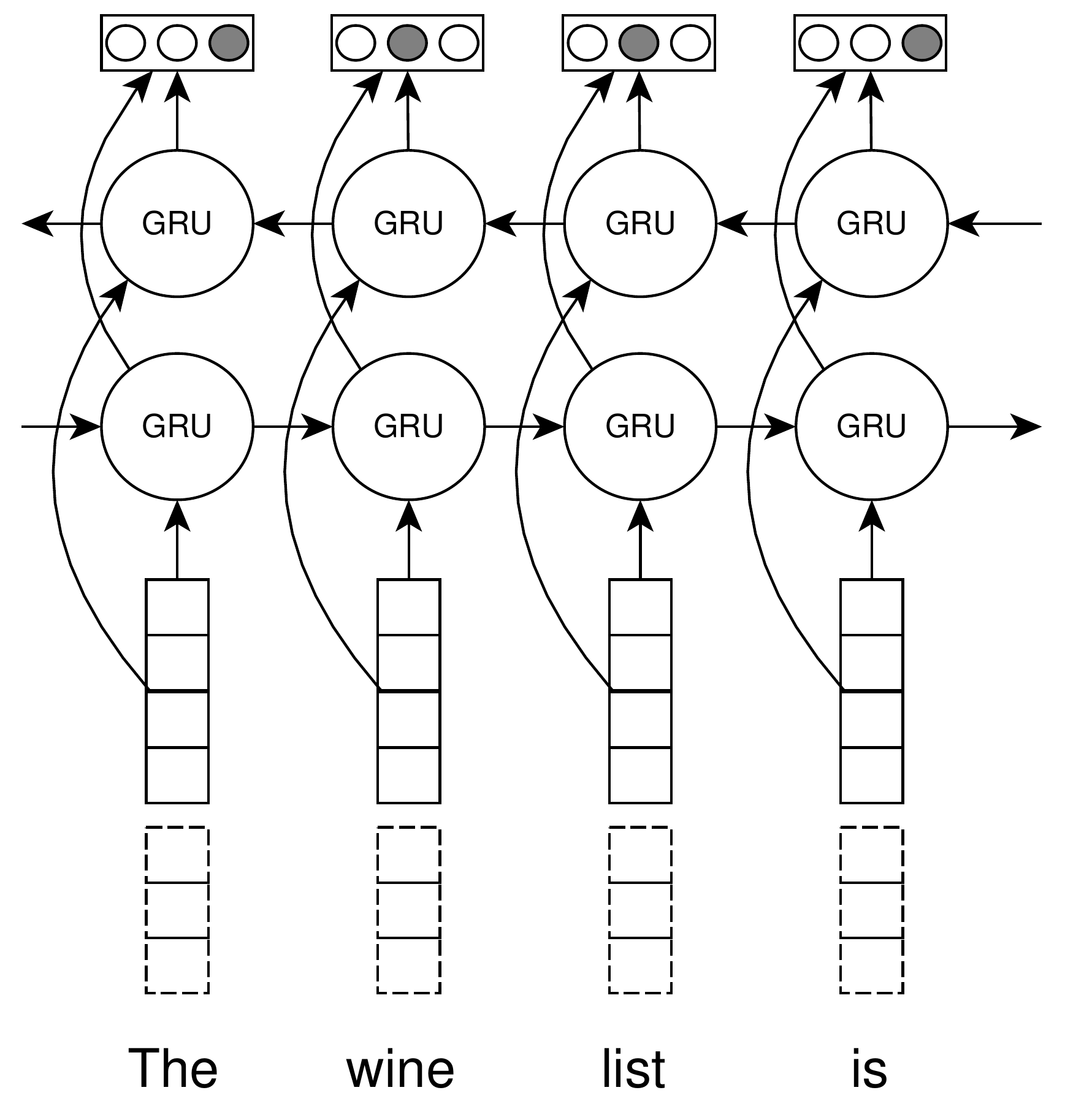}
    \caption{Illustration of the RNN sequence labeling model. The dashed boxes represent the character-level word embeddings that are only present in the character-enhanced model.}
    \label{fig:baselinemodel}
\end{figure}

The proposed baseline model is a four-layer recurrent neural network that receives a word sequence $\mathbf{w}=\{\mathbf{w}_1,\ldots,\mathbf{w}_n\}$ as input features and predicts an output sequence of IOB tags $\mathbf{t}=\{\mathbf{t}_1,\ldots, \mathbf{t}_n\}$.
Figure \ref{fig:baselinemodel} illustrates the baseline neural network.

Formally, the word sequence is passed to a word embedding layer that maps each word $\mathbf{w}_i$ to its embedding vector $\mathbf{x}^{\wrd}_i$ by means of an embedding matrix $\mathbf{W}^\wrd \in \mathbb{R}^{d^\wrd\times |V^\wrd|}$:
$$
\mathbf{x}^\wrd_i = \mathbf{W}^\wrd\mathbf{e}^{\mathbf{w}_i}
$$
where  $V^\wrd$ is the vocabulary of the word embeddings and $\mathbf{e}^{\mathbf{w}_i}$ is a one-hot vector of size $|V^\wrd|$ representing the word $\mathbf{w}_i$.

The sequence of word embedding vectors is passed to a bidirectional layer \cite{Schuster1997} of Gated Recurrent Units (GRU, \citet{cho2014learning}).
The GRU uses a combination of update and reset gates to improve its ability to learn long range information comparable to Long Short-Term Memory cells \cite{Chung2014}.
The computation of a single GRU layer at timestep\footnote{Each word in the input sequence is considered a timestep.} $i$ is as follows:

\begin{align*}
\mathbf{z}_i &= \sigma(\mathbf{W}^z \mathbf{x}_i + \mathbf{U}^z \mathbf{h}_{i-1} + \mathbf{b}^z) \\
\mathbf{r}_i &= \sigma(\mathbf{W}^r \mathbf{x}_i + \mathbf{U}^r \mathbf{h}_{i-1} + \mathbf{b}^r) \\
\mathbf{h}_i &= (1 - \mathbf{z}_i) \odot \mathbf{h}_{i-1} + \mathbf{z}_i\odot\mathbf{g}_i \\
\mathbf{g}_i &= f(\mathbf{W}^h \mathbf{x}_i + \mathbf{U}^h(\mathbf{r}_i\odot \mathbf{h}_{i-1}) + \mathbf{b}^h) 
\end{align*}
where $\mathbf{x}_i$ is an element of a generic input sequence and $\mathbf{g}_i$ the computed output. $\mathbf{z}_i$ is the update gate and $\mathbf{r}_i$ the forget gate, $\sigma$ is the sigmoid activation function and $f$ is a non-linearity for which we chose the ELU \cite{clevert2015fast} activation function.

The bidirectional GRU is a variant of the GRU that processes the input sequence in forward and backward direction.
The hidden states of the forward pass and the backward pass are concatenated to produce a single hidden state sequence:
$$
\mathbf{g} = \{[\overrightarrow{\mathbf{g}}_1:\overleftarrow{\mathbf{g}}_1], \ldots, [\overrightarrow{\mathbf{g}}_n:\overleftarrow{\mathbf{g}}_n]\}
$$
where $\overrightarrow{\mathbf{g}}_i$ and $\overleftarrow{\mathbf{g}}_i$ are the hidden states for the forward and backward GRU layer, respectively.
We choose the dimensionality of the parameters such that $\overrightarrow{\mathbf{g}}_i, \overleftarrow{\mathbf{g}}_i \in  \mathbb{R}^{r^\wrd/2}$.

The bidirectional connections allow the model to include words appearing before and after each timestep into the computation of the hidden states.
The resulting sequence of hidden states $\mathbf{g}$ presumably incorporates the necessary context for each word in its corresponding hidden state.
In a last step, each hidden state $\mathbf{g}_i$ is projected to a probability distribution $\mathbf{q}_i$ over all possible output tags, namely \textbf{I}, \textbf{O} and \textbf{B}, using a standard feedfoward layer with a softmax activation function:
$$
\mathbf{q}_i = softmax(\mathbf{W}^{tag}\mathbf{g}_i + \mathbf{b}^{tag})
$$
with $\mathbf{W}^{tag} \in \mathbb{R}^{d^{tag} \times r^\wrd}$ and $\mathbf{b}^{tag} \in \mathbb{R}^{d^{tag}}$.
For each word, we choose the tag with the highest probability as the predicted IOB tag.
The predicted tag sequence can be decoded into a set of opinion term expressions using the IOB scheme in reverse.

The trainable parameters of this model are $\mathbf{W}^\wrd$, $\mathbf{W}^{tag}$, $\mathbf{b}^{tag}$, and the parameters of the GRU $\mathbf{W}^{h}$, $\mathbf{U}^{h}$, $\mathbf{b}^{h}$, $\mathbf{W}^{z}$, $\mathbf{U}^{z}$, $\mathbf{b}^{z}$, $\mathbf{W}^{r}$, $\mathbf{U}^{r}$, $\mathbf{b}^{r}$ (for both directions).

\subsection{Character-Enhanced Model}
\label{sec:charmodel}
\begin{figure}
    \centering
    \includegraphics[width=\columnwidth]{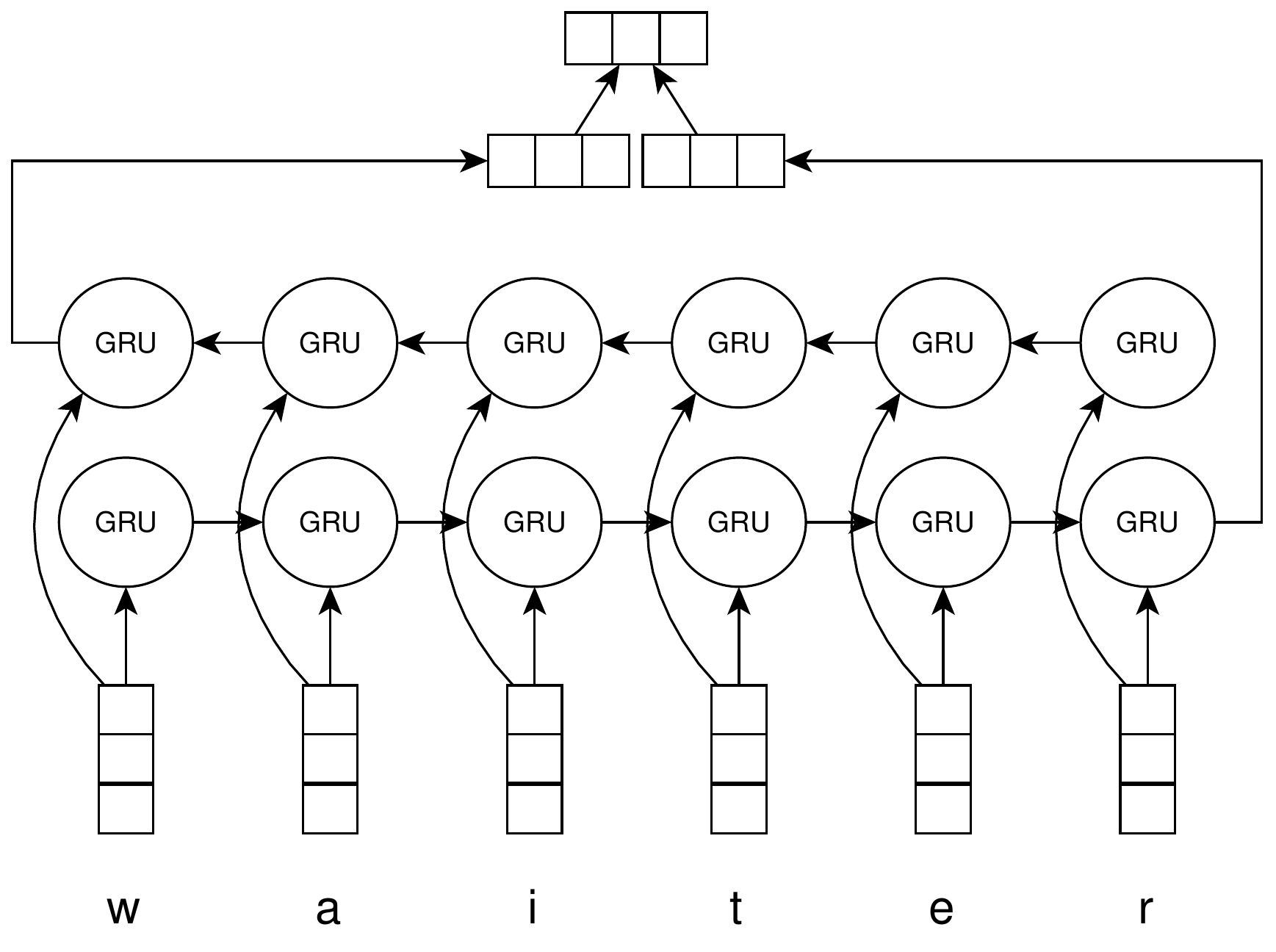}
    \caption{Illustration of the RNN word embedding model. The output of this sub network is later concatenated with the regular word embeddings.}
    \label{fig:charmodel}
\end{figure}

We propose a variation of the baseline model from Section \ref{sec:baselinemodel} that incorporates character-level information in the process of opinion target extraction.
With that we validate our hypothesis that character information poses a valuable source of information for this task.
Following previous work in this direction, we incorporate the character information in the form of character-level word embeddings.
Figure \ref{fig:charmodel} illustrates the character-level word model.
Given the character sequence $\mathbf{c} = \{c_1, \ldots, c_n\}$ of a word $w$, we first transform each character $c_i$ to its corresponding character embedding $\mathbf{x}^\chr_i$ using a character embedding matrix $\mathbf{W}^\chr \in \mathbb{R}^{d^\chr \times |V^\chr|}$:
$$
\mathbf{x}^\chr_i = \mathbf{W}^\chr\mathbf{e}^{c_i}.
$$
Analogously to the procedure for word embeddings, $V^\chr$ is the character vocabulary and $\mathbf{e}^{c_i}$ is a one-hot vector of size $|V^\chr|$ representing the character $c_i$.
As before, the sequence of character embeddings is passed through a bidirectional GRU layer that produces two sequences of hidden states, $\overrightarrow{\mathbf{g}}$ and $\overleftarrow{\mathbf{g}}$.
We choose the dimensionality of the parameters such that $\overrightarrow{\mathbf{g}}_i, \overleftarrow{\mathbf{g}}_i \in  \mathbb{R}^{d^\chr}$.

To represent the sequence of characters as a fixed-sized vector, we concatenate the final hidden states\footnote{Note that the final hidden state of the backwards directed GRU is the hidden state that corresponds to the first character in the sequence.} of both sequences and obtain a single representation $\mathbf{g}=[\overrightarrow{\mathbf{g}_n}: \overleftarrow{\mathbf{g}_1}]$ for the character sequence.
Lastly, the concatenated hidden state $\mathbf{g}$ is transformed to the final character-level word embedding using a linear feedforward layer:
$$
\mathbf{x}^{cw} = \mathbf{W}^{cw}\mathbf{g} + \mathbf{b}^{cw}
$$
with $\mathbf{W}^{cw} \in \mathbb{R}^{d^\chr \times 2\cdot d^\chr}$ and $\mathbf{b}^{cw} \in \mathbb{R}^{d^\chr}$.

To incorporate the word model in the overall neural network model, we pass the corresponding character sequence of each word in $\mathbf{w}=\{\mathbf{w}_1,\ldots, \mathbf{w}_n\}$ through the word model to obtain $\mathbf{x}^{cw}=\{\mathbf{x}^{cw}_1,\ldots, \mathbf{x}^{cw}\}$.
The resulting character-level embeddings are then concatenated with the word level embeddings:
$$
\mathbf{\widetilde{x}} =\{[\mathbf{x}^{w}_1:\mathbf{x}^{cw}_1], \ldots, [\mathbf{x}^{w}_n:\mathbf{x}^{cw}_n]\}
$$
The augmented sequence $\mathbf{\widetilde{x}}$ replaces $\mathbf{x}$ in the baseline model and is passed through the remaining layers of the network.
Since $\mathbf{\widetilde{x}}$ contains word and character-level information, the following RNN and projection layers can theoretically make use of the additional information to improve the extraction of opinion target expressions.

The trainable parameters of this model are $\mathbf{W}^{w}$,$\mathbf{W}^{c}$,$\mathbf{W}^{cw}$, $\mathbf{b}^{cw}$, $\mathbf{W}^{tag}$, $\mathbf{b}^{tag}$, and the parameters of the GRU $\mathbf{W}^{h}$, $\mathbf{U}^{h}$, $\mathbf{b}^{h}$, $\mathbf{W}^{z}$, $\mathbf{U}^{z}$, $\mathbf{b}^{z}$, $\mathbf{W}^{r}$, $\mathbf{U}^{r}$, $\mathbf{b}^{r}$ for the word and character-level RNN (and for both directions).

\subsection{Network Training}
The optimization of the model parameters is done by minimizing the classification error for each word in the sequence using the cross-entropy loss.
The optimization is carried out using a mini-batch size of 5 with the stochastic optimization technique \emph{Adam} \cite{Kingma2014}.
We clip the norm of the gradients to 5 and regularize our network quite rigorously using L2 regularization of $10^{-5}$ on $\mathbf{W}^{tag}$ and $\mathbf{W}^{cw}$ and Dropout in various positions in our network.
Initial experiments suggested that this regularization is necessary due to the moderate size of the training dataset.
The networks are implemented using the machine learning framework \emph{Keras} \cite{keras2015}.

The word embedding matrix $\mathbf{W}^\wrd$ is initialized with a pretrained matrix of skip-gram embeddings trained on a corpus of amazon reviews \cite{McAPanLes15}.
Earlier work showed that using a domain specific corpus in the pretraining stage significantly improves performance for similar tasks \cite{jebbara2016aspect}.

\section{Experiments and Evaluation}
\label{sec:experiments}
In this section, we evaluate the impact of using character-level word embeddings on the task of extracting opinion target expressions from user-generated reviews.
For this, we compare the character-enhanced model from Section \ref{sec:charmodel} to the baseline RNN of Section \ref{sec:baselinemodel}.
We start by describing the used dataset in Section \ref{sec:dataset}.
To select a fitting set of hyperparameters for each model, we perform a 5-fold cross validation on the training portion of our dataset.
Using the best hyperparameters, we evaluate both models on the test portion of the data and investigate the models' properties with respect to the induced character information in Sections \ref{sec:testresults} and \ref{sec:analysis}.
Evaluation is carried out in terms of F1 score of gold opinion target expressions and retrieved opinion term expressions using exact matches\footnote{We use the provided evaluation code from the organizers of the SemEval 2016 challenge.}.

\subsection{Dataset}
\label{sec:dataset}
In this work, we use the dataset from previous SemEval Tasks to evaluate our assumption.
Specifically, we use the data for the aspect-based sentiment analysis challenge of the year 2016 (Task 5).
The used dataset consists of review sentences from the restaurant domain with annotations for opinion target expressions.
Table \ref{tab:datasets} gives a summary of the dataset.
\begin{table}[t]
    \centering
    \begin{tabular}{l|ccc}
        \hline
        Dataset & \#Sent. & \#OTEs & \#Chars per OTE\\
        \hline
        Train & 2000 & 1880 & 2 -80 \\
        Test & 676 & 650 & 3-50  \\
        \hline
    \end{tabular}
    \caption{Relevant statistics of the SemEval 2016 dataset (Task5, restaurant domain).}
    \label{tab:datasets}
\end{table}

\subsection{Hyperparameter Selection}
\label{sec:parametersearch}
We set the dimensionality $d^\wrd$ of the pretrained word embeddings to 100 and perform a grid search on a subset of the hyperparameters to find a suitable solution to be used in the final analysis.
We evaluate each candidate set of hyperparameters using a 5-fold cross validation on the training data.
The search is performed for each model (\texttt{word-only} and \texttt{char+word}).
We experiment with:
\begin{itemize}
\item the size of the word vocabulary\footnote{The size of the word vocabulary is the main factor in terms of (GPU) memory usage.} $|V^\wrd| \in \{10000, 20000, 50000\}$ (with respect to the most frequent words),
\item the size of the sentence level RNN hidden layer $r^\wrd \in \{60, 100, 200\}$,
\item and the size of the character-level RNN and the corresponding character-level word embedding vector $d^\chr \in \{20, 50, 100\}$.
\end{itemize}

Table \ref{tab:hyperparameters} shows the best hyperparameters for each model.
\begin{table}[t]
    \centering
    \begin{tabular}{l|ccccc}
        \hline
         Model & $|V^\wrd|$ & $r^\wrd$ & $d^\chr$ & $\varnothing$ F1 \\
        \hline
         \texttt{word-only} & 50000 & 60 & -- & 0.6713\\
         \texttt{char+word} & 50000 & 100 & 100 & \textbf{0.6936}\\ 
        \hline
    \end{tabular}
    \caption{Results of a search for hyperparameters. The column \textit{$\varnothing$ F1-Score} gives the best mean score for the best performing training epoch across cross validation models.}
    \label{tab:hyperparameters}
\end{table}
As expected, the search indicates that it is always better to increase the size of the word vocabulary $V^\wrd$.
The best model using both word and character-level information performs on average about 2.2 points F1-Score better than the best model that only uses word-level information.
For the following evaluations, we instantiate and train our models according to these hyperparameters.

\subsection{Results on Test}
\label{sec:testresults}
For the evaluation on the test set, we use the previously found hyperparameters and instantiate our models.
We train both models on 80\% of the training set and use the remaining 20\% as a validation set for early stopping \cite{CaruanaLG00}.
The \texttt{word-only} model reaches its best performance at epoch 35 and the \texttt{char+word} model peaks at epoch 73.

The performances of both models are given in Table \ref{tab:testresults}.
The results confirm our hypothesis and the findings from the cross validation that the character-level word embeddings offer a substantial improvement (3.3 points F1-Score ) over the \texttt{word-only} baseline model.
\begin{table}[t]
    \centering
    \begin{tabular}{l|c}
        \hline
        Model & F1-Score \\
        \hline
        \texttt{word-only} &  0.6260\\
        \texttt{char+word} & \textbf{0.6586}\\ 
        \hline
    \end{tabular}
    \caption{Results on the test set for best performing hyperparameters. The previous findings of the usefulness of character-level word embeddings are confirmed by the results of the test set.}
    \label{tab:testresults}
\end{table}

\subsection{Analysis}
\label{sec:analysis}
In this Section, we investigate what the character-level word embeddings encode and if there are specific cases in which the character-enhanced model performs better than the baseline.

\begin{figure*}[t]
\centering
\begin{subfigure}{.5\textwidth}
  \centering
  \includegraphics[width=\columnwidth]{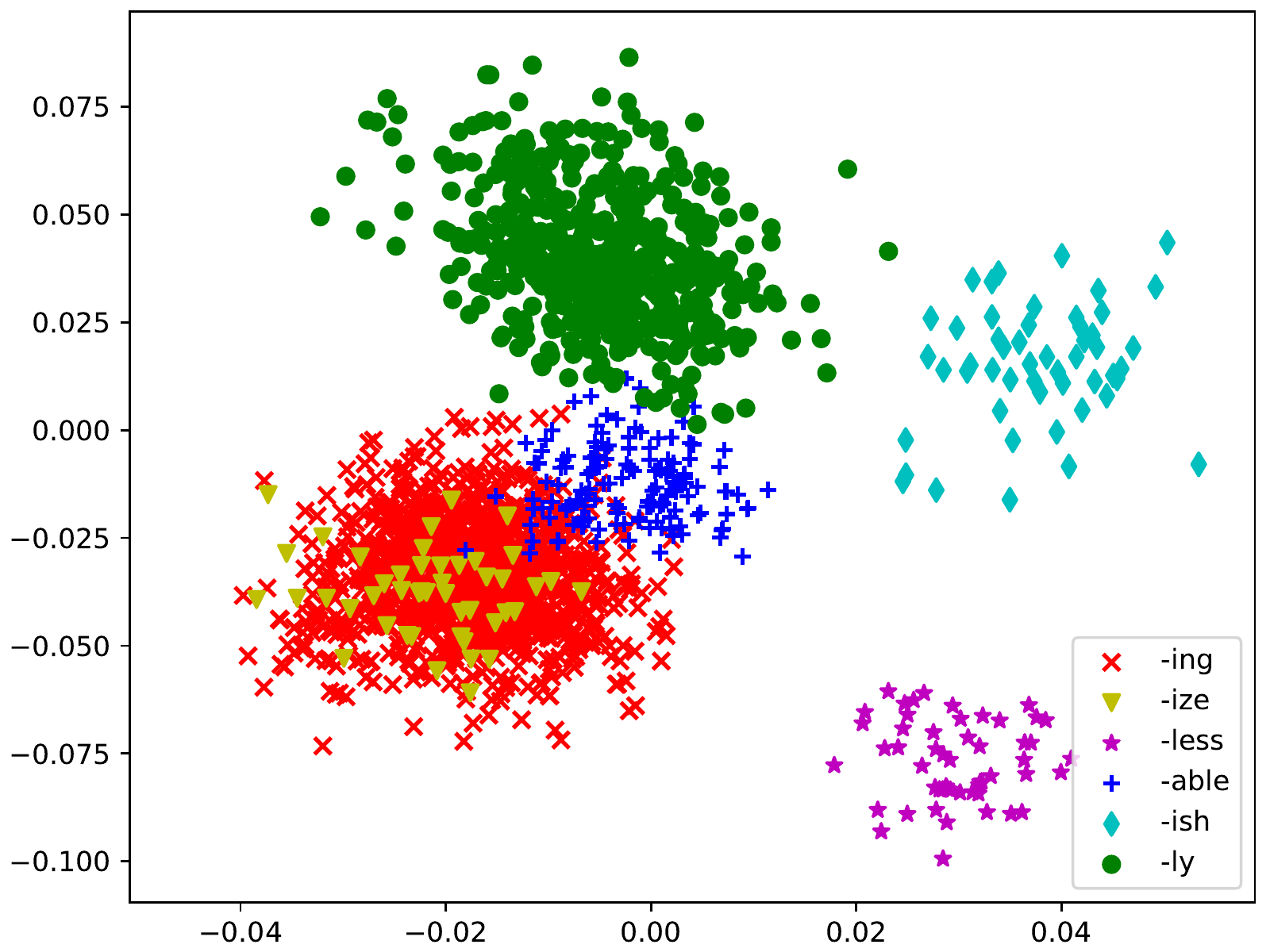}
  \caption{Character-level embeddings}
  \label{fig:charscatter}
\end{subfigure}%
\begin{subfigure}{.5\textwidth}
  \centering
  \includegraphics[width=\columnwidth]{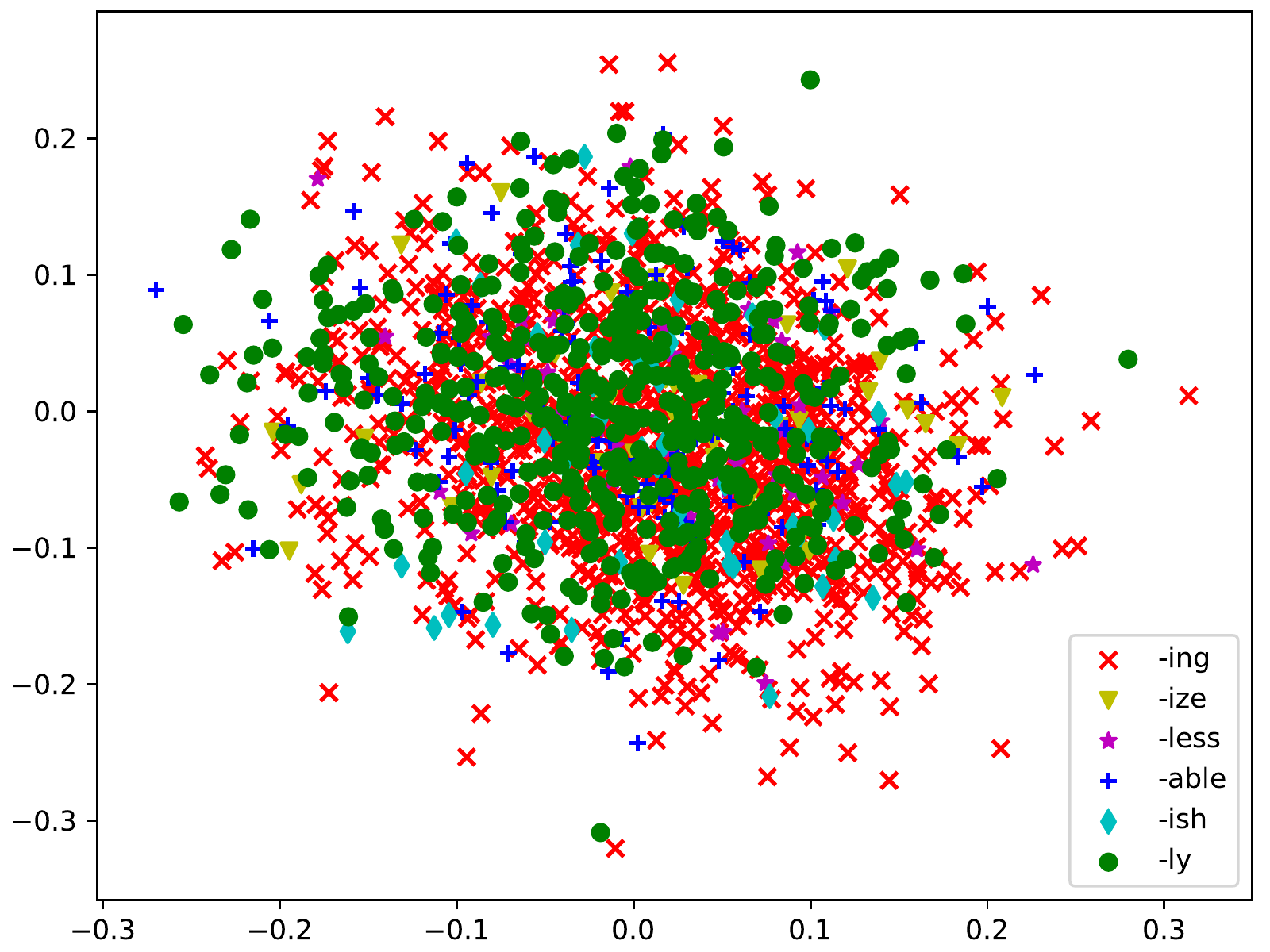}
  \caption{Word-level embeddings}
  \label{fig:wordscatter}
\end{subfigure}
\caption{Visualization of suffix information of the two employed types of embeddings.}
\label{fig:scatter}
\end{figure*}

\paragraph{Visualization}
Our initial experiments in visualizing the learned model suggested that the character-level word embeddings encode morphological features of a word.
To confirm this assumption, we visualize the learned embeddings using suffix information.
We extract a subset of the 2000 most frequently occurring words from reviews that end on one of the following suffixes: -ing, -ly, -able, -ish, -less, -ize.
We project the embeddings of the words to a 2 dimensional space using T-SNE \cite{tsne2008} and plot them as a scatter plot.
By highlighting each word according to its suffix, we see that the character-level embeddings are grouped according to their suffixes (see Figure \ref{fig:charscatter}).
Performing the same procedure with the regular skip-gram word embeddings results in no clear separation between the 6 suffix groups (see Figure \ref{fig:wordscatter}).

Previous work in the direction of aspect-based sentiment analysis shows a positive impact of POS tag features for the extraction of aspect terms and opinion target expressions \cite{Toh2014,jebbara2016aspect}.
It stands to reason if the character-level word embeddings act in a similar way.
The morphological information of character-level word embeddings (as shown in Figure \ref{fig:charscatter}) might help to disambiguate word occurrences with respect to their linguistic function in the sentence, similar to the positive effect of POS tags for this task.
We leave this hypothesis for future work.

\paragraph{Out-of-Vocabulary Errors}
Next, we are interested in seeing if the improvement in F1-score can be backtraced to Out-of-Vocabulary (OOV) word errors.
For this, we compute the F1-score on 3 different subsets of sentences for the \texttt{word-only} model and the \texttt{char+word} model:
\begin{itemize}
\item \texttt{no OOV}: This subset only contains sentences for which all words are part of the known vocabulary.
\item \texttt{OOV sent.}: This subset contains sentences that contain an unknown word at some position in the sentence.
\item \texttt{OOV op.}: The subset of sentences that contain at least one opinion target expression with an unknown word.
\end{itemize}

\begin{figure*}[t]
\centering
\begin{subfigure}{.5\textwidth}
  \centering
  \includegraphics[width=\columnwidth]{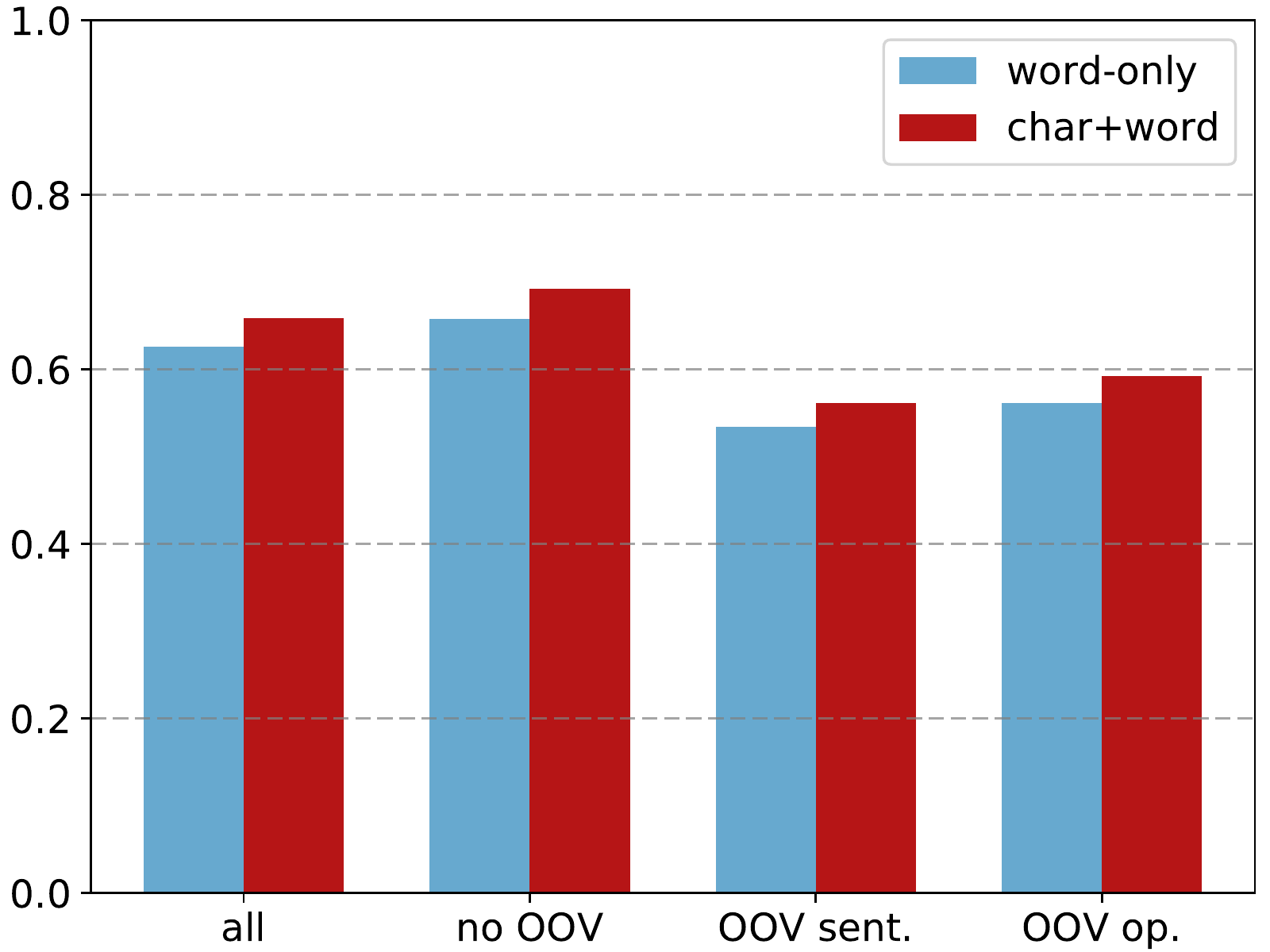}
  \caption{Performance on OOV-related subsets.}
  \label{fig:oov}
\end{subfigure}%
\begin{subfigure}{.5\textwidth}
  \centering
  \includegraphics[width=\columnwidth]{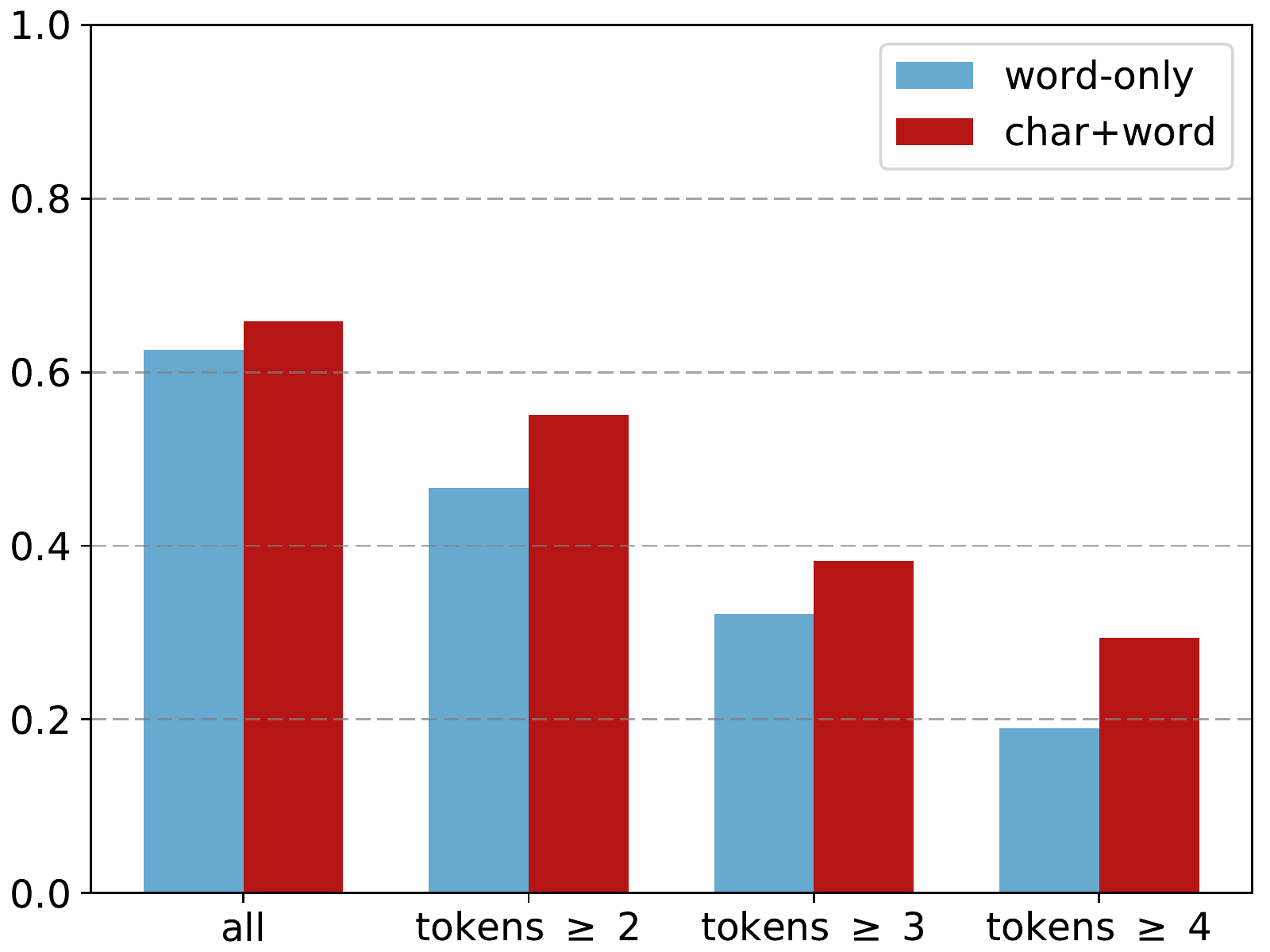}
  \caption{Performance on multi word phrases.}
  \label{fig:multiword}
\end{subfigure}
\caption{Illustration of performance differences for different subsets of sentences.}
\label{fig:subsets}
\end{figure*}

Figure \ref{fig:oov} shows F1-scores for different subsets.
Surprisingly, we can see that the F1-scores rise and fall similarly for both models regardless of the evaluated subset.
This suggests, that the positive influence of the character information does not particularly help in those cases where the text contains previously unseen words (e.g.\ misspelled words).
We assume that the positive impact on these cases is mitigated since the domain specific skip-gram word embeddings already contain various writing errors that frequently occur in customer reviews.
This can be seen in Table \ref{tab:misspelled} that shows the nearest neighbors of exemplary words in the skip-gram embedding space.
We see that common writing mistakes are often already captured by the word embeddings.
\begin{table*}[t]
\centering
\begin{tabular}{lccc}
\toprule
\multicolumn{1}{l}{Word}            & atmosphere  & restaurant  & service  \\
\midrule
\multirow{3}{*}{Nearest Neighbors} & \textit{atomosphere}  & \textit{restaraunt} & customer \\
				  & ambience & eatery & \textit{serivce}   \\
				  & \textit{atmoshpere} & \textit{restuarant} & costumer\\
\bottomrule
\end{tabular}
\caption{Three commonly used words in restaurant reviews and their 3 nearest neighbors in the embedding space. Often, misspelled versions (\textit{italic}) of the original word are among its closest neighbors.} 
\label{tab:misspelled}
\end{table*}

\paragraph{Multi-Word Expressions}
Another possible cause for the performance difference of both models might be related to the length of opinion target expressions\footnote{In terms of words.}.
This hypothesis is motivated by the idea that e.g.\ variations in spelling with respect to hyphenation (e.g.\ \textit{bartenders} vs. \textit{bar tenders} or \textit{wait staff} vs. \textit{wait-staff}) could have less of an influence on the character-based model than on the word-based model.
To test this idea, we consider subsets of sentences that contain at least one OTE that is a multi-word expression of more than or equal to $k$ words.
The performance differences for $k \in \{2,3,4\}$ are visualized in Figure \ref{fig:multiword}.

The first thing to notice is that both models are strongly effected by the length of the OTEs.
Longer expressions seem to be harder to extract in general.
However, we can observe that the character model is influenced by the length of an OTE to a lesser degree.
While the difference in F1-score for all sentence between the \texttt{word-only} model and \texttt{char+word} model is about 3.3, the differences for OTEs composed of more than or equal to 2, 3, and 4 words are 8.4, 6.1 and 10.4, respectively.

\section{Conclusion}
\label{sec:conclusion}
There is a growing interest in character and subword-level models for natural language processing in recent years.
Tokenization is a crucial step for many applications, yet neglects the information that can be gained from the character structure of a word itself.

In this work, we were able to show that character-level information assists in the task of opinion target extraction, an important step in aspect-based sentiment analysis.
We compared a model using only word-level features to a more sophisticated model that also includes character-level word embeddings.
We showed that the more complex character model consistently outperforms the baseline model with a substantial margin of 3.3 points F1-score.
A visualization of the learned embeddings revealed encoded morphological regularities that we could not find in our skip-gram word embeddings.
Through experiments on different subsets of the data, we linked the positive influence of the character-level word embeddings to the difficulty of extracting multi-word expressions.
We did not observe a performance difference for Out-of-Vocabulary cases.

However, it is not entirely clear how exactly the additional character information contributes to the task of extracting opinion target expression.
In general, we suspect that the morphological information of character-level word embeddings helps to disambiguate word occurrences similar to the positive effect of POS tags for OTE extraction.
A confirmation of this hypothesis remains for future work.

Another interesting direction for future work is the pretraining of parts of the network to enrich the character-based word representation.
We believe that character-level language models pose an interesting candidate for this.

The positive results of this work and the remaining research questions suggest a need to focus further research effort in this direction in order to improve token-based approaches or even replace the need for tokenization altogether.

\section*{Acknowledgments}
This research/work was supported by the Cluster of Excellence Cognitive Interaction Technology 'CITEC' (EXC 277) at Bielefeld University, which is funded by the German Research Foundation (DFG).

\bibliography{emnlp2017}
\bibliographystyle{emnlp_natbib}

\end{document}